\documentclass[letterpaper, 10 pt, conference]{ieeeconf}

\IEEEoverridecommandlockouts
\usepackage{graphics} % for pdf, bitmapped graphics files
\usepackage{epsfig} % for postscript graphics files
\usepackage{times} % assumes new font selection scheme installed
\usepackage{amsmath} % assumes amsmath package installed
\usepackage{amssymb}  % assumes amsmath package installed
\usepackage[table]{xcolor}
\usepackage[ruled,linesnumbered, noend]{algorithm2e}
\usepackage{flushend} % Balance column in last page

\usepackage{soul}
\usepackage{cite}
\usepackage{comment}
\usepackage{multirow}
\usepackage{graphicx}
\usepackage{wrapfig}
\usepackage{soul, color}
\usepackage{wrapfig}
\usepackage{mathtools}
\usepackage{graphicx}
\usepackage{booktabs}
\usepackage{adjustbox}

\usepackage{caption}
\usepackage{subfigure}

\definecolor{LightCyan}{rgb}{0.88,1,1}
\definecolor{Gray}{gray}{0.9}

\newcommand{\ve}[1]{\mathbf{#1}} % vector
\newcommand{\hve}[1]{\hat{\mathbf{#1}}} % vector
\newcommand{\tve}[1]{\tilde{\mathbf{#1}}} % vector
\newcommand{\tg}{SightGAN} 
\newcommand{\ts}{\tg~} 
\newcommand{\sr}{sim-to-real } 
\newcommand{\rs}{real-to-sim } 

\title{Augmenting Tactile Simulators with Real-like and Zero-Shot Capabilities}

\author{Osher Azulay$^*$ Alon Mizrahi$^*$, Nimrod Curtis$^*$ and Avishai Sintov
\thanks{$^*$ These authors contributed equally.}
\thanks{O. Azulay, A. Mizrahi, N. Curtis and A. Sintov are with the School of Mechanical Engineering, Tel-Aviv University, Israel. E-mail: \{osherazulay, alonmizrahi2, nimrodcurtis\}@mail.tau.ac.il; sintov1@tauex.tau.ac.il.}
}

\begin{document}

\setlength{\belowdisplayskip}{2pt}
\setlength{\belowdisplayshortskip}{3pt}
\setlength{\abovedisplayskip}{2pt} 
\setlength{\abovedisplayshortskip}{3pt}
\setlength{\parskip}{0pt}

% \markboth{IEEE Robotics and Automation Letters. Preprint Version. Accepted January, 2019}
% {Sintov \MakeLowercase{\textit{et al.}}: Learning a State Transition Model of an Underactuated Adaptive Hand}

\maketitle
\thispagestyle{empty}
\pagestyle{empty}

% \begin{IEEEkeywords}
%   Tendon/Wire Mechanism, Underactuated Robots, Dexterous Manipulation.
% \end{IEEEkeywords}

\begin{abstract}
    Simulating tactile perception could potentially leverage the learning capabilities of robotic systems in manipulation tasks. However, the reality gap of simulators for high-resolution tactile sensors remains large. Models trained on simulated data often fail in zero-shot inference and require fine-tuning with real data. In addition, work on high-resolution sensors commonly focus on ones with flat surfaces while 3D round sensors are essential for dexterous manipulation. In this paper, we propose a bi-directional Generative Adversarial Network (GAN) termed \tg. \ts relies on the early CycleGAN while including two additional loss components aimed to accurately reconstruct background and contact patterns including small contact traces. The proposed \ts learns \rs and \sr processes over difference images. It is shown to generate real-like synthetic images while maintaining accurate contact positioning. The generated images can be used to train zero-shot models for newly fabricated sensors. Consequently, the resulted \sr generator could be built on top of the tactile simulator to provide a real-world framework. Potentially, the framework can be used to train, for instance, reinforcement learning policies of manipulation tasks. The proposed model is verified in extensive experiments with test data collected from real sensors and also shown to maintain embedded force information within the tactile images.
\end{abstract}

\section{Introduction}

Tactile sensing is a fundamental aspect of human perception and, therefore, is a topic for extensive research in robotics \cite{xu2023efficient,church2022tactile,suresh2023midastouch}. Such sensing plays a crucial role in enabling robots to interact with the physical world potentially with precision and dexterity. The advancement of tactile sensor technologies has led to high-dimensional and complex data representations which, in turn, requires sufficient data in order to train accurate models and policies \cite{Dong2021}. Hence, tactile simulations and \sr approaches are gaining momentum as researchers explore new ways to bridge the gap between virtual and physical worlds \cite{higuera2023learning,gomes2023beyond}. 
%%%%%%%%%%%%%%%%%%%%%%%%%%%
\begin{figure}
    \centering
    \includegraphics[width=0.9\linewidth]{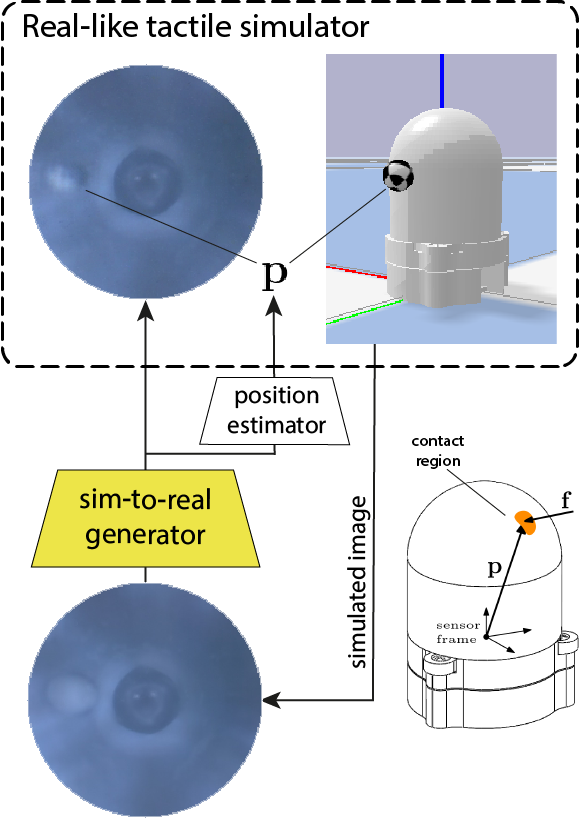}
    \caption{\small The \sr generator from the trained \ts model is used to map simulated tactile images to real-like images of a 3D round tactile sensor. Since the generated image is close to reality, various models can be trained using the simulator. In this example, a position estimator can provide accurate labeling to the image making it a fine simulator for various tasks.}
    \label{fig:tacto}
    \vspace{-0.6cm}
\end{figure}
%%%%%%%%%%%%%%%%%%%%%%%%%%%
%%%%%%%%%%%%%%%%%%%%%%%%%%%%
\begin{figure*}[h]
    \centering
    \includegraphics[width=\linewidth]{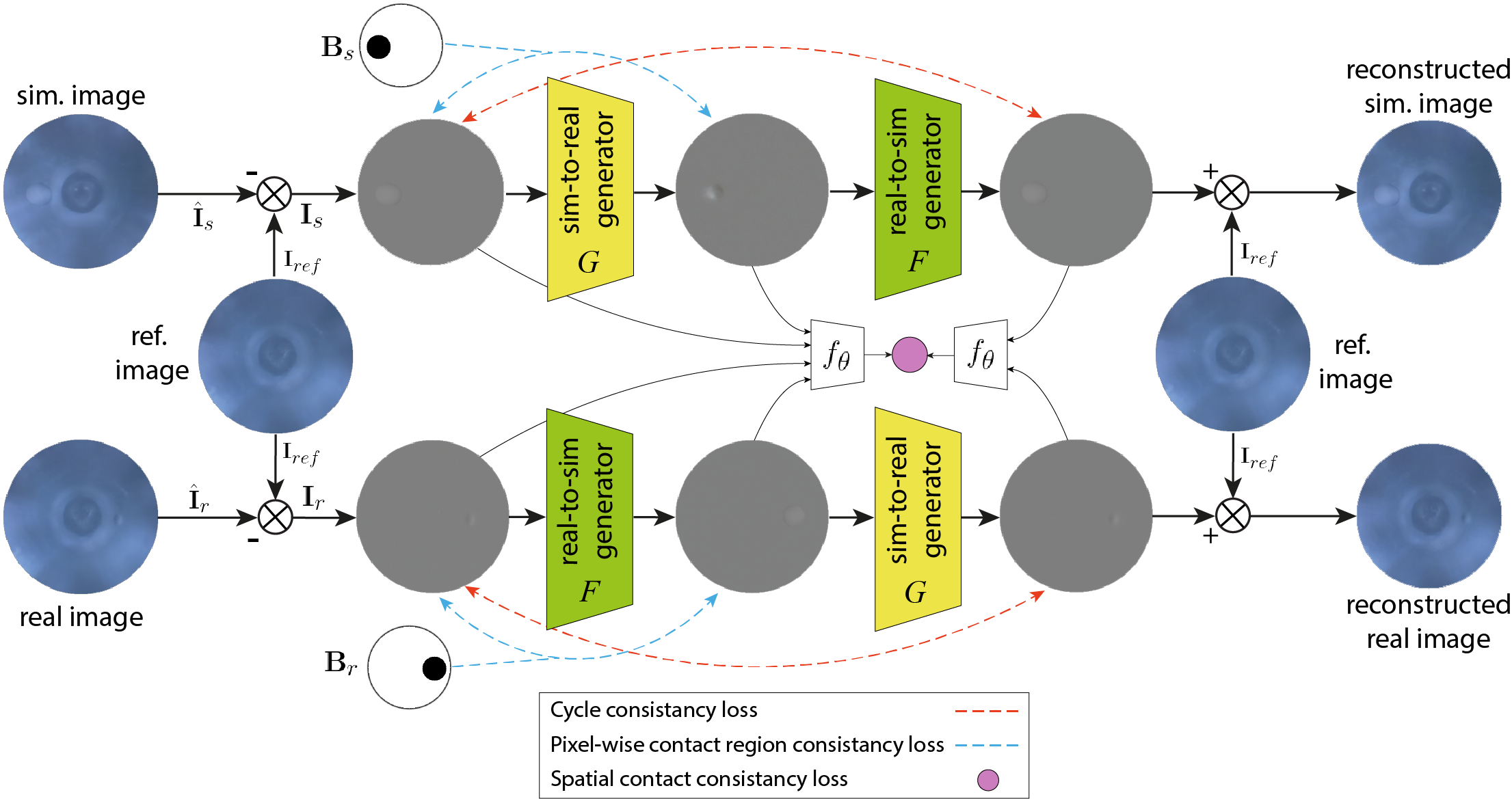}
    \caption{\small Scheme of the \ts model. The model operates on difference images in order to enhance generability to new sensors. Top and bottom rows illustrate the sim-to-real-to-sim and real-to-sim-to-real processes, respectively. The cycle consistency loss of CycleGAN is augmented by two additional losses aimed to provide pixel-level domain adaptation of the contacts.}
    \label{fig:scheme}
    \vspace{-0.6cm}
\end{figure*}
%%%%%%%%%%%%%%%%%%%%%%%%%%%%

% This endeavor is particularly relevant in robotics \cite{higuera2023learning, gomes2023beyond, xu2023efficient,church2022tactile, suresh2023midastouch}, where accurate representation and translation of tactile data are essential for enhancing performance and interaction.

Tactile sensors come in various technologies including capacitive transducers \cite{maslyczyk2017highly}, force sensitive resistors \cite{flx2022finger} and piezo-resistors \cite{cheng2009novel}. However, they usually fit to specific applications and provide low-resolution data. Optical-based tactile sensors, on the other hand, have become increasingly common due to their ability to provide high-resolution signals \cite{taylor2022gelslim, lambeta2020digit, yuan2017gelsight,azulay2023allsight}. In such sensors, an internal camera observes the deformation of the contact pad, typically made of a soft elastomer, during contact with an object. An image captured by the internal camera has the potential to encode essential information regarding the contact including its position with respect to the sensor's coordinate frame. Nevertheless, due to the rich data in such images, training models to estimate features in the data requires an extensive amount of samples.

In order to cope with the data amount requirements, simulations of optical-based tactile sensors have been addressed and have the potential to rapidly generate large datasets of tactile images \cite{wang2022tacto, si2022taxim}. However, transferring a model trained on simulated data to a real sensor, i.e., sim-to-real, may present difficulties. Primarily, real-world tactile images often exhibit substantial disparities when compared to their simulated counterparts \cite{patel2020deep, si2022grasp}. 
% The intricate task of modeling the optical reactions of gel deformations and integrating the dynamics of contact interactions poses challenges in the simulation of vision-based tactile sensors. 
In attempt to bridge the gap between simulated tactile data and real-world tactile information, various approaches have been explored. Domain randomization was included in simulation of the TacTip sensor where some parameters are constantly randomised \cite{Ding2020}. Similarly, a simulation was augmented with random texture perturbations in order to train a GelSight sensor model \cite{Fernandes2021}. Domain randomization, however, requires careful choices of the parameters to randomize and is limited in complex tasks such as in high-resolution tactile images. Some work have harnessed Finite Element Methods (FEM) in order to generate simulated deformation of the contact pad in tactile \sr \cite{Narang2021}. However, the computational complexity of such an approach limits real-time sensing \cite{Sferrazza2022}.

A notable technique involves the use of Generative Adversarial Networks (GANs), a class of deep learning models that can generate new image samples from learned data distributions \cite{goodfellow2014}. In tactile real-to-sim, a GAN was used to match real images of the BioTac sensor in order to target simulated ones \cite{church2022tactile}. Yet, the approach requires paired images which are difficult to extract. Hence, a variant of GAN termed CycleGAN \cite{zhu2017unpaired} has gained traction for its ability to facilitate \sr and \rs transfer of tactile information without a paired dataset. 
% With its ability to perform unsupervised domain adaptation, is well-suited for such situations. 
CycleGAN is able to learn the mapping of data distributions from the simulated source domain to the real-world target domain and vice versa, without explicit correspondence between individual samples. 
% This capacity to bridge the gap between different data domains makes CycleGAN-alike methods an invaluable tool for sim2real learning in tactile sensing. The concept of sim2real entails learning from simulated data and transferring that knowledge to real-world scenarios, a process that is integral to achieving robust and adaptable robotic interactions.
% https://ieeexplore.ieee.org/stamp/stamp.jsp?tp=&arnumber=10106009&tag=1
% Several studies have leveraged the power of GANs to address the challenges of tactile sensor sim2real transfer. 
In the first work to utilize CycleGAN in tactile sensing, a bidirectional \sr approach was proposed for the GelSight sensor \cite{chen2022bidirectional}. In a later work, an improved CycleGAN architecture was introduced with task-specific loss functions for enhanced structural fidelity of generated tactile images \cite{jing2023unsupervised}. However, these approaches and others \cite{zhao2023skill,kim2023marker} focused on a specific tactile sensor with a flat contact surface and without exhibiting zero-shot capability. In addition, prior work focused on tactile images where the contact trace seen in the image is rather large \cite{xu2023efficient}. Consequently, CycleGAN may learn to hide information in the adapted image instead of explicitly retaining the semantics when the trace is small \cite{rao2020rl,chu2017cyclegan}.

% Chen et al. \cite{chen2022bidirectional} proposed a bidirectional sim2real transfer approach for GelSight tactile sensors using CycleGAN, effectively bridging the gap between simulated and real data. Jing et al. \cite{jing2023unsupervised} introduced an improved CycleGAN architecture with task-specific loss functions for enhanced structural fidelity of generated tactile images, promoting more accurate sim2real transfer. Moreover, Zhao et al. \cite{zhao2023skill} focused on object in-hand orientation estimation using CycleGAN, incorporating a contact region consistency loss to highlight specific tactile features crucial for the task. Kim et al. \cite{kim2023marker} devised a marker-embedded tactile GAN approach, aligning simulated depth with real pressing to enable learned depth-to-marker RGB generation. 

In this paper, we tackle the \sr problem for high-resolution 3D round sensors while enabling zero-shot inference of accurate contact position estimation (Figure \ref{fig:tacto}). 
Building upon CycleGAN, we propose the \textit{\tg} model which augments CycleGAN with contact-specific consistency losses as illustrated in Figure \ref{fig:scheme}. The losses reduce background disparities between simulated and real tactile images while minimizing contact position errors. For the latter, distillation with a trained contact position estimator is exerted to compare accuracies between generated tactile images, either in real or simulated domains. One of the key advantages of employing \ts is its bidirectional capability, allowing to seamlessly transfer knowledge both from real to simulated domains and vice versa. This versatility enables to train models for various sensors of different illumination and fabrication uncertainties.

\ts is evaluated on the novel AllSight sensor \cite{azulay2023allsight} whereas it can potentially be applied to any optical-based tactile sensor. AllSight is a high-resolution and all-round tactile sensor in which a model for it requires a sufficient amount of data. Hence, our approach provides a simulated environment with real-like and accurately labeled tactile images as demonstrated in Figure \ref{fig:tacto}. As such, models trained on these synthetic images exhibit a zero-shot inference capability on new, real and untrained sensors. In addition, unlike prior work, \ts is able to withstand small contact traces. Hence, the proposed model can be used in applications with small loads in, for instance, in-hand manipulation. The simulator of AllSight along with the proposed \ts \sr framework and datasets are provided open-source\footnote{AllSight simulator, \ts \sr framework and datasets: \texttt{https://github.com/osheraz/allsight\_sim}} for the benefit of the community and to advance research in the field.

\section{Optical tactile sensors}
\label{sec:optical}

\subsection{Design}
\begin{figure}
    \centering
    \includegraphics[width=\linewidth]{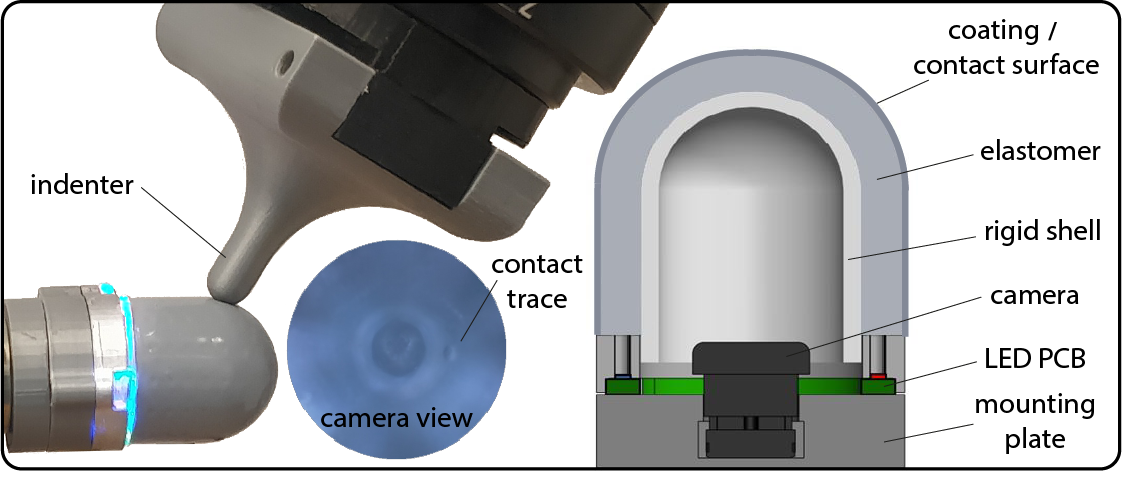}
    \caption{\small (Left) The AllSight tactile sensor with the internal view of the camera during contact with a round indenter. (Right) Structure illustration of the AllSight sensor.}
    \label{fig:allsight}
    \vspace{-0.7cm}
\end{figure}

An optical-based tactile sensor typically uses an internal camera to track the deformation of a soft elastomer upon contact with an object. In this work, we consider the AllSight sensor proposed in \cite{azulay2023allsight} which has a full 360$^\circ$ round contact region clearly visible without blind spots or obscurance. Hence, the camera is covered by a tube in the shape of a cylinder with an hemispherical end as seen in Figure \ref{fig:allsight}. The tube is three-layered where the inner layer is a rigid crystal-clear shell. A transparent elastomer covers the shell and is coated on its exterior by a reflective silicone paint. In such design, the camera observes the deformation of the elastomer from within upon contact with the exterior layer. The inner-surface of the shell is evenly illuminated by an annular printed circuit board (PCB) with embedded LEDs. The lighting provides better visibility and informative images. In this work we consider white lighting while any other LED setting is possible.
%%%%%%%%%%%%%%%%%%%%%%%%%%%
%%%%%%%%%%%%%%%%%%%%%%%%%%%
% %%%%%%%%%%%%%%%%%%%%%%%%%%%
% \begin{figure}
%     \centering
%     \includegraphics[width=\linewidth]{Figures/allsight.png}
%     \caption{\small Image of the simulation environment and the camera view.}
%     \label{fig:tacto}
% \end{figure}
% %%%%%%%%%%%%%%%%%%%%%%%%%%%

\subsection{Data collection of real images}
\label{sec:real_collection}

A dataset $\mathcal{P}_{real}$ of real images with spatial position labels is collected using an automated setup. A robotic arm with a round indenter mounted on its tip repeatedly touched the surface of AllSight in various contact locations. First, a reference image $\ve{I}_{ref}\in\mathcal{I}_r$, where $\mathcal{I}_r$ denotes the space of real images, is recorded for a sensor without any contact. During contact, an image $\hve{I}_i$ is taken along with its position $\ve{p}_i$ on the contact surface. The position is represented in the sensors coordinate frame as seen in Figure \ref{fig:tacto} and calculated through the forward kinematics of the arm. Furthermore, we consider difference images in the dataset such that an image used for training is $\ve{I}_i=\hve{I}_i-\ve{I}_{ref}$. Such subtraction has been shown to enhance the learning process making the model agnostic to the background and focuses only on the color gradients that occur around the deformations \cite{higuera2023learning}. Consequently, the model is expected to generalize to new sensors of different backgrounds in zero-shot. The acquisition and labeling process yields dataset $\mathcal{P}_{real}=\{(\ve{I}_i,\ve{p}_i) \}_{i=1}^N$ of $N$ labeled images. 

\subsection{Data collection in simulation}
\label{sec:sim_collection}

TACTO is a physics-engine simulator for optical-based tactile sensors \cite{wang2022tacto}. An AllSight simulation was set in TACTO as seen in Figure \ref{fig:tacto} and calibrated by including reference images from real AllSight sensors. To enhance sim-to-real pre-training of the state estimation model, we collected different reference images from different AllSight sensors and used them for augmentation. %The acquired images were augmented by adding noise and varying the lighting conditions. 
The simulated dataset $\mathcal{P}_{sim}$ was generated by labeling $M$ images captured in TACTO during random contacts. Here also, a reference image from a real sensor is subtracted from the contact images. An image $\ve{I}_i\in\mathcal{I}_s$, where $\mathcal{I}_s$ denotes the space of simulated images, is taken along with the contact position $\ve{p}_i$ such that $\mathcal{P}_{sim}=\{(\ve{I}_i,\ve{p}_i) \}_{i=1}^M$. 

\subsection{Contact position estimation model}

Using either datasets $\mathcal{P}_{real}$ or $\mathcal{P}_{sim}$, we train a contact position estimation model. The model $f_\theta:\mathcal{I}\to\mathbb{R}^3$ maps a tactile image to the spatial position of the contact on the sensor $\ve{p}\in\mathbb{R}^3$. Vector $\theta$ is the trainable parameters of the model. The model is based on the ResNet-18 architecture \cite{he2016deep}. The top layer is removed and the flattened output features are fed through two fully-connected layers of size 512 and 256. At each iteration, both reference $\ve{I}_{ref}$ and contact $\ve{I}_i$ images are down-sampled to resolution $224\times224$ and stacked along the channel. The stacked image is then passed through the model to get the estimated position $\tve{p}_i$.

\section{Method}

The proposed \tg, illustrated in Figure \ref{fig:scheme}, integrates the CycleGAN architecture with additional auxiliary losses designated for tactile images. The losses aim to reduce disparities in background and contact reconstruction of images in the bidirectional transfer between simulation and real domains. 
% These are underpinned by the recognition that disparities primarily reside in color and illumination aspects, which have been effectively tackled by CycleGAN in diverse applications \cite{ho2021retinagan}. 
In this section, we briefly outline the GAN and CycleGAN losses prior to presenting the \ts loss.

% The additional losses aim bridge the gap between simulated and real tactile images. 

% --------------------------------------

\subsection{Generative Adversarial Network (GAN)}

In GAN, a mapping $G:\mathcal{I}_s\to\mathcal{I}_r$ is trained such that a discriminator $D_r$ cannot distinguish between an original image $\ve{I}_r\in\mathcal{I}_r$ and a synthetic one $\tve{I}_r=G(\ve{I}_s)$ where $\ve{I}_s\in\mathcal{I}_s$. Model $G$ is trained to minimize the adversarial loss 
\begin{align}
    \label{eq:gan}
    \mathcal{L}_{\text{GAN}}(G,D_r,\mathcal{I}_s,\mathcal{I}_r)&=\mathbb{E}_{\ve{I}_r \sim \mu_r} [\log(D_r(\ve{I}_r))]\\ 
    & +\mathbb{E}_{\ve{I}_s \sim \mu_s} [\log(1 - D_r(G(\ve{I}_s)))] \nonumber
\end{align}
where $\mu_r$ and $\mu_s$ are the data distributions of $\mathcal{I}_r$ and $\mathcal{I}_s$, respectively.

% --------------------------------------

\subsection{CycleGAN: Bidirectional Image Translation}

Our approach facilitates bidirectional mapping, as in CycleGAN, between unpaired image datasets from two domains, $\mathcal{I}_{s}$ and $\mathcal{I}_{r}$. This process involves the generators of $G: \mathcal{I}_{s} \rightarrow \mathcal{I}_{r}$ and $F: \mathcal{I}_{r} \rightarrow \mathcal{I}_{s}$ along with adversarial discriminators $D_s$ and $D_r$. In additional to adversarial losses for both mappings, a cycle consistency loss is included in order to minimize the cyclic reconstruction of images given by 
\begin{align}
    \mathcal{L}_{\text{cycle}}(G,D_r,\mathcal{I}_s,\mathcal{I}_r)&=\mathbb{E}_{\ve{I}_r \sim \mu_r} \| 
G(F(\ve{I}_r))-\ve{I}_r\|_1\\ 
    & +\mathbb{E}_{\ve{I}_s \sim \mu_s} \| F(G(\ve{I}_s))-\ve{I}_s\|_1 \nonumber.
\end{align}
% $\mathcal{F}(\mathcal{G}({\ve{I}_{s}})) \approx {\ve{I}_{s}} $ and $\mathcal{G}(\mathcal{F}({\ve{I}_{r}})) \approx {\ve{I}_{r}}$ for $\ve{I}_{s}\in\mathcal{I}_{s}$ and $\ve{I}_{r}\in\mathcal{I}_{r}$, respectively. 
Consequently, CycleGAN is trained with the following loss:
\begin{align}
\label{eq:CycleGAN}
\mathcal{L}_{\text{CycleGAN}}(G, F, D_s, D_r) &= \mathcal{L}_{\text{GAN}}(G, D_r, \mathcal{I}_{s}, \mathcal{I}_{r})  \\
& + \mathcal{L}_{\text{GAN}}(F, D_s, \mathcal{I}_{r}, \mathcal{I}_{s}) \nonumber \\
& + \lambda_{\text{cycle}}\mathcal{L}_{\text{cycle}}(F,G) \nonumber
\end{align}
where $\lambda_{\text{cycle}}$ is some pre-defined weight.

% --------------------------------------

\subsection{SightGAN Losses}% Auxiliary Losses}

Drawing inspiration from the perception consistency loss of RetinaGAN \cite{ho2021retinagan}, we introduce two auxiliary contact losses, one in the image domain and the other in the contact space. In the context of optical tactile images, these images can be partitioned into two distinct regions: the background region representing the no-contact area of the tactile sensor and the foreground region embodying the tactile sensor's interaction with objects. For the background region, the \textit{Pixel-wise Contact Region Consistency loss} emphasizes the constraint of color similarity in the generated images while forcing no contact traces in the region. Furthermore, the \textit{Spatial Contact Consistency Loss} focuses on the contact estimation accuracy as minor deviations in the generated image can have great significance on the spatial positioning of the contact. This meticulous attention to foreground texture and structural nuances of the background is pivotal in optimizing \sr tactile images from 3D tactile sensors. 

\subsubsection{Spatial Contact Consistency loss}

The spatial contact consistency loss penalizes disparities in contact localization after transferring images across domains. We define function $\mathcal{L}_{sp}$ to compare between contact position estimations of two images, $\ve{I}$ and $\ve{J}$ as
%the original image and the transformed one given by 
\begin{equation}
    \mathcal{L}_{sp}(\ve{I},\ve{J}) = \|f_\theta(\ve{I})-f_\theta(\ve{J})\|^2.
\end{equation}
% where $H$ is either $F$ or $G$ for real or simulated images, respectively. 
With this function and based on the perception consistency loss of the RetinaGAN \cite{ho2021retinagan}, we define the spatial contact consistency loss as
\begin{align}
\label{eq:spatial_loss}
\mathcal{L}_{\text{spatial}}({\ve{I}_{s}}, {\ve{I}_{r}}, F, G) &= \mathcal{L}_{sp}(\ve{I}_s,  G(\ve{I}_{s}))+ \\
+ \frac{1}{2} \mathcal{L}_{sp}(\ve{I}_s,  F(G(\ve{I}_s))) &+ \frac{1}{2} \mathcal{L}_{sp}(G(\ve{I}_s),  F(G(\ve{I}_s)))+ \nonumber \\
+ \mathcal{L}_{sp}(\ve{I}_r,  F(\ve{I}_r)) &+ \frac{1}{2} \mathcal{L}_{sp}(\ve{I}_r, G( F(\ve{I}_r)))+ \nonumber \\
& + \frac{1}{2} \mathcal{L}_{sp}( F(\ve{I}_r), G( F(\ve{I}_r))) \nonumber .
\end{align}
The halving of losses involving the cycled images accounts for their dual comparison against the original and transferred images.

\subsubsection{Pixel-wise Contact Region Consistency loss}

In order to further augment the the accuracy of contact localization in domain transfer and enhance structural fidelity, we introduce a loss related to the contact region. 
Minor changes in contact pixels can lead to significant alterations in the contact domain, potentially causing errors in contact estimation with $f_\theta$. 
For either simulated or real training images, the contact position $\ve{p}$ is labeled as described in Section \ref{sec:optical}. For each image $\ve{I}$, binary image $\ve{B}$ is defined where a mask is placed on the contact region of the image. In practice, pixels in and out of the contact regions are marked with zeros and ones, respectively, in $\ve{B}$. The contact loss between an image and its transfer is, therefore, defined by
\begin{equation}
    \mathcal{L}_{\text{m}}(\ve{I},\ve{B},H) = \|\ve{I}\ast\ve{B} - H(\ve{I})\ast\ve{B}\|_1
\end{equation}
where $\ast$ denotes pixel-wise multiplication and, $H$ is either $F$ or $G$ for real or simulated images, respectively. Hence, the pixel-wise contact region consistency loss is defined as
\begin{equation}
    \label{eq:mask_loss}
    \mathcal{L}_{\text{mask}}(\ve{I}_s,\ve{I}_r,F,G) = \mathcal{L}_{\text{m}}(\ve{I}_s,\ve{B}_s,G)+\mathcal{L}_{\text{m}}(\ve{I}_r,\ve{B}_r,F).
\end{equation}
Note that additional loss components in \eqref{eq:mask_loss} similar to $\mathcal{L}_{\text{spatial}}$ do not enhance performance as resulted in preliminary analysis and, therefore, not included.

% Given masks $r_i$ representing pixels within the observed contact regions for simulated ($x$) and real ($y$) images, we define the following loss with masks $\ve{m}_{\ve{I}_{s}}$ and $\ve{m}_{\ve{I}_{r}}$:
% \begin{equation}
% \mathcal{L}_{mask} = \mathcal{L}{1}(m_{\ve{I}}\cdot{\ve{I}}, m_{\ve{I}}\cdot\mathcal{G}(\ve{I}))
% \end{equation}

% \begin{align}
% \mathcal{L}_{c_i}({\ve{I}_{s}}, {\ve{I}_{r}}, F, G) &= \mathcal{L}_{aux_i}({\ve{I}_{s}},  (G(\ve{I}_{s})) + \frac{1}{2} \mathcal{L}_{aux_i}({\ve{I}_{s}},  F(G({\ve{I}_{s}}))) \nonumber \\
% & + \frac{1}{2} \mathcal{L}_{aux_i}(G({\ve{I}_{s}}),  F(G(\ve{I}_{s}))) \nonumber \\
% & + \mathcal{L}_{aux_i}({\ve{I}_{r}},  F({\ve{I}_{r}})) + \frac{1}{2} \mathcal{L}_{aux_i}({\ve{I}_{r}}, G( F({\ve{I}_{r}}))) \nonumber \\
% & + \frac{1}{2} \mathcal{L}_{aux_i}( F({\ve{I}_{r}}), G( F({\ve{I}_{r}})))
% \end{align}
% \begin{equation}
%     \mathcal{L}_{c}({\ve{I}_{s}}, {\ve{I}_{r}}, \mathcal{F}, \mathcal{G}) = \mathcal{L}_{sptl}({\ve{I}_{s}}, {\ve{I}_{r}}, \mathcal{F}, \mathcal{G}) + \mathcal{L}_{mask}({\ve{I}_{s}}, {\ve{I}_{r}}, \mathcal{F}, \mathcal{G})
% \end{equation}

% \subsection{\ts loss}

The above two auxiliary losses are applied across batches of simulated and real images with sim-to-real $G$ and real-to-sim $F$ generators. Hence, the overall auxiliary loss is defined by the sum of both
\begin{align}
    \label{eq:aux}
    \mathcal{L}_{c}({\ve{I}_{s}}, {\ve{I}_{r}}, F, G) &= \lambda_{\text{spatial}}\mathcal{L}_{\text{spatial}}({\ve{I}_{s}}, {\ve{I}_{r}}, F, G) \\
    & + \lambda_{\text{mask}}\mathcal{L}_{\text{mask}}({\ve{I}_{s}}, {\ve{I}_{r}}, F, G) \nonumber
\end{align}
where $\lambda_{\text{spatial}}$ and $\lambda_{\text{mask}}$ are weight parameters. Combining \eqref{eq:CycleGAN} with \eqref{eq:aux}, the complete \ts loss is 
\begin{align}
\mathcal{L}_{\text{\tg}}(G, F, D_s, D_r) &= \mathcal{L}_{\text{GAN}}(G, D_r, \mathcal{I}_{s}, \mathcal{I}_{r}) \nonumber \\
&\quad + \mathcal{L}_{\text{GAN}}(F, D_s, \mathcal{I}_{r}, \mathcal{I}_{s}) \nonumber \\
&\quad + \lambda_{\text{cycle}}\mathcal{L}_{\text{cycle}}(F, G) \nonumber \\
&\quad +  \mathcal{L}_{c}({\ve{I}_{s}}, {\ve{I}_{r}}, F, G).
\end{align}

Once the \ts model has been trained, a simulator is available. In order to generate real-like images for a new sensor, one only needs to reintegrate the generated foreground image outputted from the \sr generator with a real reference image of the sensor. Hence, the generated images are expected to provide zero-shot contact inference.

% The overall approach combines these foundational principles with additional losses specific to CycleGAN, forming a cohesive framework for sim2real transfer and real2sim adaptation, as depicted in Figure X. This integration empowers the model to simultaneously learn from real-world tactile data and insights intrinsic to simulated data, enhancing the fidelity of synthesized images by effectively retaining background color intricacies and foreground contact textures.

\section{Experiments}

Experiments are presented in this section in order to analyze the performance of \tg. While prior work often use one specific sensor, we collect a training dataset $\mathcal{P}_{real}$ over six different AllSight sensors. The dataset is collected in an automated process where each AllSight sensor is mounted on a fixed frame. An indenter with a round tip of radius 3 mm is mounted to a robotic arm. The arm is also equipped with a Robotiq FT-300 Force/Torque (F/T) sensor for labeling contact forces. 
% The system is controlled using the Robot Operating System (ROS). During the collection, data stream is acquired in a frequency of 60 Hz. 
For each sample, the robot selects a point to press on the surface of the sensor and computes its location through its kinematics. An image taken during the press is labeled with its position. The collection yielded 1,000 labeled images for each sensor and a total of $N=6,000$ labeled images in $\mathcal{P}_{real}$. In addition, we aim to analyze the generalization abilities of the trained model for new sensors. Hence, a test set is collected over two new sensors not included in the training. The test set is comprised of 2,000 labeled images. Results are presented with white LED illumination for all sensors whereas RGB illumination was tested in preliminary experiments yielding similar results. Also, for each image, the reference image is subtracted as described in Section \ref{sec:real_collection}.

For $\mathcal{P}_{sim}$, data is collected in TACTO as described in Section \ref{sec:sim_collection}. $\mathcal{P}_{sim}$ consists of simulated tactile images and their associated contact positions, featuring a 3 mm radius spherical indenter at varying penetration depths. To fine-tune the simulation, we utilized reference images obtained from six different sensors. %Additionally, we introduced Gaussian noise and augmented the simulation with different illumination settings to enhance data diversity.
The collection yielded 1,000 labeled images for each sensor and a total of $M=6,000$ labeled images in $\mathcal{P}_{sim}$

\subsection{Contact position evaluation}

Table \ref{tb:genralization_diff_exp} summarizes the Root-Mean-Square-Errors (RMSE) for position estimation with model $f_\theta$ while trained with different origins of training data. All models were evaluated on distinct data from the two test sensors. The results include the lower accuracy bound of directly training with real data (different from the test data) from the two test sensors. Also, we include accuracy when training $f_\theta$ directly with $\mathcal{P}_{real}$. Next, model $f_\theta$ is trained with data $\mathcal{P}_{sim}$ generated in the simulation without any GAN and while using the reference images of the six training sensors. Using only simulated data provides poor accuracy showing that the simulation, even with real reference images, is far from representing reality. Then, $f_\theta$ with data generated by the CycleGAN (trained with $\lambda_\text{cycle}=10$) alone without additional losses is evaluated. The error with only CycleGAN is the highest due to its inability to focus and reconstruct the contact. This is a known issue when most of the image is not under contact and can even lead to a mode of collapse \cite{rao2020rl}. Adding either spatial contact consistency loss \eqref{eq:spatial_loss} with $\lambda_\text{spatial}=0.1$ or pixel-wise contact region consistency loss \eqref{eq:mask_loss} with $\lambda_\text{mask}=30$ is shown to significantly reduce the error. The accuracy of \tg, which combines CycleGAN with both losses, provides the lowest error out of the generative models. Hence, \ts generates real-like images from simulated ones and enables zero-shot position estimation of contacts.

\begin{figure}[h]
    \centering
    \includegraphics[width=\linewidth]{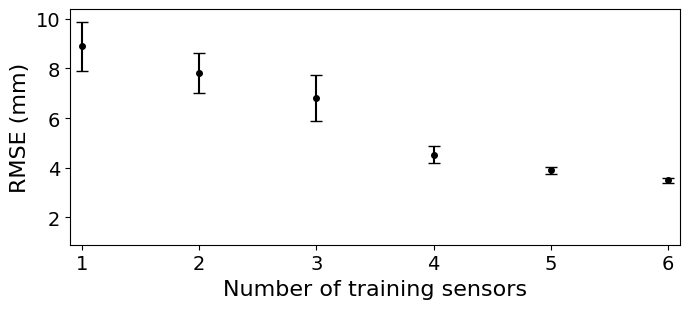}
    \caption{\small Position estimation error of $f_\theta$ trained with synthetic data from \ts over the test sensor with regards to the number of train sensors used to train \tg.}
    \label{fig:num_sensors}
    \vspace{-0.5cm}
\end{figure}

\begin{figure}[h]
    \centering
    \includegraphics[width=\linewidth]{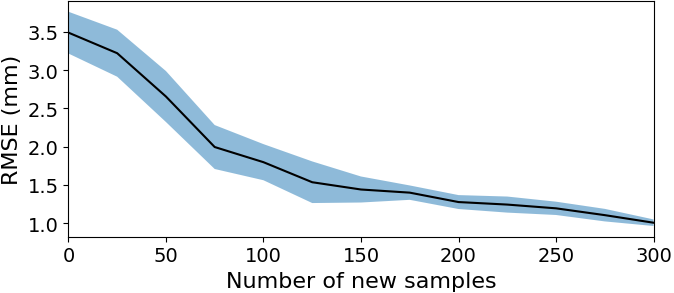}
    \caption{\small Position estimation error with regards to the number of real images from the test sensor used to fine-tune model $f_\theta$. Results with zero new tactile images are the zero-shot transfer errors without any fine-tuning.}
    \label{fig:ex3}
\end{figure}

We now evaluate the accuracy which the \sr of \ts provides to images generated from simulation considering diversity in the training data. Model $f_\theta$ is trained over synthetic images generated by the \sr of \tg. Figure \ref{fig:num_sensors} presents the position estimation RMSE
of $f_\theta$ over the test sensors with regards to the number of real train sensors used to train \tg. The addition of more sensors in the training set increases diversity and decreases the estimation error. Hence, more train sensors improves the zero-shot capability over new real sensors. While zero-shot provides relatively accurate predictions, real data from the target sensors can improve accuracy. Figure \ref{fig:ex3} shows the error of position estimation over the test data of the two new sensors with regards to the number of new samples used to fine-tune the model. The addition of a small amount of new samples for fine-tuning further improves accuracy. With 300 additional samples, the position RMSE reaches to approximately 1 mm. 

Next, we evaluate the generability of a model, trained on samples with round indenters, to estimate contacts of other geometries. 
% These images are used to train a contact position estimator $f_\theta$. 
Test data of 2,000 labeled images was collected from one test sensor with square (edge length
6 mm) and elliptical (axis lengths 8 mm and 4 mm) indenters. Table \ref{tb:diff_ind_exp} presents the RMSE results for position contact estimation on the test data. The first row shows a baseline for training and testing model $f_\theta$ with the new test indenters. All other results in the Table are based on training data of round indenters and evaluation on the test indenters. While the second and third rows are estimation models trained on real data, the rows below are based on estimation models trained with simulated images taken from the simulator while using similar square and elliptical indenters. The simulated images are passed through the respective \sr model and used to train $f_\theta$. Here also, \ts exhibits the lowest error over all \sr generated image origins. Furthermore, the results show that using \ts generalizes well and shows similar accuracy to directly training with real data. Hence, synthetic data from the \sr generator of \ts generalizes well and can be used to train zero-shot estimation models.

%%%%%%%%%%%%%%%%%%%%%%%%%%%%%%%%
\begin{table}[]
\centering
\caption{\small Estimation accuracy of contact positions over data from two test sensors with regards to the origin of the train data for $f_\theta$}
% Regression task results - Training on sensor 15,17 real, simulated, and GAN-generated Difference images - Test on real sensor 15,17}
\label{tb:genralization_diff_exp}
\begin{tabular}{clc}
\toprule
& Origin of training data for $f_\theta$                & Position RMSE (mm) \\
\midrule
\multirow{3}{*}{\rotatebox[origin=c]{90}{Direct}} & Data from the 2 test sensors       & 0.66 \\
& Data from 6 train sensors          & 2.16 \\
& 6 sensors from simulation          & 7.48 \\
\midrule
\multirow{4}{*}{\rotatebox[origin=c]{90}{Generative}} & CycleGAN                             & 13.30\\
& CycleGAN + spatial contact loss \eqref{eq:spatial_loss}      & 3.86 \\ 
& CycleGAN + pixel-wise contact loss \eqref{eq:mask_loss}   & 3.70 \\
& \ts                                  & 3.49 \\
\bottomrule
\end{tabular}
% \vspace{-0.5cm}
\end{table}
%%%%%%%%%%%%%%%%%%%%%%%%%%%%%%%%
%%%%%%%%%%%%%%%%%%%%%%%%%%%%%%%%
%%%%%%%%%%%%%%%%%%%%%%%%%%%%%%%%
%%%%%%%%%%%%%%%%%%%%%%%%%%%%%%%%
%%%%%%%%%%%%%%%%%%%%%%%%%%%%%%%%
%%%%%%%%%%%%%%%%%%%%%%%%%%%%%%%%
\begin{table}[]
\centering
\caption{\small Estimation accuracy of contact positions over data from one test sensor with indenters of different geometries}
% \caption{Regression task results - Training on sensor 15 real, simulated, and GAN-generated Difference images - Test on real sensor 15 different indenters}
\label{tb:diff_ind_exp}
\begin{tabular}{clc}
\toprule
& \multirow{2}{*}{Origin of training data for $f_\theta$} & Position \\
& & RMSE (mm) \\
\midrule
\multirow{4}{*}{\rotatebox[origin=c]{90}{Direct}} & Train sensors w/ square and ellipse indenters                   & 0.94  \\
& Train sensors                                                   & 2.67  \\
& Test sensor                                                     & 3.19  \\
& 6 simulated sensors                                             & 7.74  \\     
\midrule
% Diff mask distil cycle gan                                    & 3.883 \\
% Diff mask distil cycle gan (diff reg)                         & 3.891 \\ 
\multirow{4}{*}{\rotatebox[origin=c]{90}{Generative}} & CycleGAN                                                        & 13.28 \\
& CycleGAN + spatial contact loss \eqref{eq:spatial_loss}         & 3.93  \\ 
& CycleGAN + pixel-wise contact loss \eqref{eq:mask_loss}         & 3.97  \\
& \ts                                                             & 3.67 \\
\bottomrule
\end{tabular}
% \vspace{-0.5cm}
\end{table}
%%%%%%%%%%%%%%%%%%%%%%%%%%%%%%%%

\subsection{Sim-to-real quality}

In order to evaluate the \sr of \ts for generating realistic tactile images, we consider the Frechet Inception Distance (FID) \cite{Heusel2017} and Kernel Inception Distance (KID) \cite{Binkowski2018} metrics. FID and KID are common pixel-level metrics of domain adaptation and quantify the quality and diversity of generated images. Furthermore, they do not require paired images. Table \ref{tb:fid_kid} exhibits FID and KID comparative results for direct mapping between simulated and real images, for CycleGAN and for \tg. Direct comparison between simulated and real images is obviously poor and emphasizes the reality gap. CycleGAN provides some improvement. Nevertheless, \ts is shown to be superior to CycleGAN with approximately 47\% and 16\% quality improvements for FID and KID, respectively.

%%%%%%%%%%%%%%%%%%%%%%%%%%%%%%%%
\begin{table}[]
    \centering
    \caption{\small FID and KID comparison over the generated tactile images.}
    \label{tb:fid_kid}
    \begin{tabular}{lccc}
        \toprule
        Model && FID & KID \\
        \midrule
        Direct   & $\ve{I}_r$ vs. $\ve{I}_s$    & 98.54$\pm$0.90 & 29.19$\pm$0.45 \\
        CycleGAN & $\ve{I}_r$ vs. $G(\ve{I}_s)$ & 73.63$\pm$0.51 & 13.27$\pm$0.10 \\
        \tg      & $\ve{I}_r$ vs. $G(\ve{I}_s)$ & 39.12$\pm0.30$ & 11.24$\pm$0.12 \\
         \bottomrule
    \end{tabular}
    \vspace{-0.5cm}
\end{table}
%%%%%%%%%%%%%%%%%%%%%%%%%%%%%%%%

\subsection{Contact force evaluation}

Common tactile simulations provide force readings which are linear to the deformation of the contact pad \cite{wang2022tacto,Si2022}. In reality, however, modeling the deflection of the contact pad is nearly impossible and the reaction forces are more complex \cite{Ruppel2019,Narang2021}. Hence, we further assess the potential of using \ts to provide accurate force estimations to the real-like simulated tactile images. A trained contact force estimation model $g_\psi:\mathcal{I}\to\mathbb{R}^3$ is evaluated to approximate the corresponding contact force $\ve{f}=[f_x,f_y,f_z]^T\in\mathbb{R}^3$. Vector $\psi$ includes trainable model parameters. A real dataset was collected with the F/T sensor by labeling tactile images of the test sensors with force magnitudes during contact. The dataset includes a wide range of contact force magnitudes within ranges $f_z\in[0N, 12N]$ and $f_x,f_y\in[-2N, 2N]$ where $f_z$ is the normal force and $f_x,f_y$ are the forces tangential to the contact surface at $\ve{p}$.

Table \ref{tb:force_exp} presents a comparison of force estimation accuracy over several origins of train data for $g_\psi$. For a baseline, model $g_\psi$ is trained on data from the six train sensors and evaluated on different data from the same sensors. Similarly, the model was trained on data from the two test sensors and evaluated on different data from the same sensors. Obviously, having more diversity originating from more sensors provides better generability and lower errors. Next, \ts and $g_\psi$ are trained on data from the six training sensors. Following, force accuracy is evaluated by passing new test data, either from the six train sensors or the two test sensors, through the \rs and \sr generators. The reconstructed images are then evaluated with model $g_\psi$. 

The results with \ts show relatively similar accuracy to the baseline errors. Evidently, \ts maintains force information embedded within the original images and reconstructs it. Potentially, simulated images from the simulator upon contact can be mapped with the \sr to the real domain and evaluated with $g_\psi$. Hence, the simulator would be able to provide a comprehensive contact state in various applications. However, such approach requires further research and evaluations. We leave this for future work.

% , we evaluated a trained contact force model $g_\theta(\ve{I})$ to identify the corresponding contact force $\ve{f}\in\mathbb{R}^3$ during pressing actions. To achieve this, we equipped the robotic arm with a Force/Torque (F/T) sensor and gathered a dataset consisting of $x$ images capturing various contact interactions with their corresponding force using the AllSight sensor. The dataset included a wide range of contact force magnitudes, within the following specified limits: $f_z\in[-12N, 0.8N]$, while $f_x$ and $f_y$ fell within the range of $[-5N, 5N]$. We note that as the simulation environment lacks the capability to offer accurate force estimations, for labeling we matched our real dataset by passing it through both the real-to-sim generator and the \sr generator once more. Table \ref{tb:force_exp} summarizes the force estimation errors for model $g_\theta$ while trained with different origins of training data. The results include the lower accuracy bound of directly training with real data from the two test sensors. Results show quite accurate estimation for both train and test sensors, with less then \hl{update numbers}.  

%%%%%%%%%%%%%%%%%%%%%%%%%%%%%%%%
\begin{table}[]
\centering
\caption{\small Estimation accuracy of contact forces with regards to the origin of the train data for $g_\psi$}
\label{tb:force_exp}
\begin{tabular}{llc}
\toprule
% \multicolumn{1}{c|}{data}  & \multicolumn{4}{c}{\textbf{RMSE (N)}}  \\ \hline
Origin of train data for $g_\psi$   & Test data$^1$   & Force RMSE (N) \\
\midrule
6 train sensors & 2 train sensors         & 0.30 \\ 
2 test sensors  & 2 test sensors               & 0.78 \\ 
\ts real-to-sim-to-real & 2 train sensors & 0.42 \\ 
\ts real-to-sim-to-real & 2 test sensors       & 0.81 \\ 
\bottomrule
\multicolumn{3}{l}{\footnotesize $^1$Train and test data are distinct in all cases.} \\
\vspace{-0.6cm}
\end{tabular}
\end{table}

\section{Conclusions}

In this paper, a novel generative model termed \ts was proposed for augmenting tactile simulators with real-like capabilities. The model is based on a bidirectional GAN and can be used to simulate 3D round tactile sensors. Using unpaired real and simulated data, \ts is trained to provide \sr and \rs mappings while minimizing the cycle consistency loss of CycleGAN and two additional loss components. The two loss components aim to minimize contact localization errors in the reconstruction of the image through distillation with a trained contact position estimator. Such approach augments a simulator with real-like performance and enables zero-shot capabilities for models trained on the simulated sensors and evaluated on real ones. Indeed, using real data to train a contact position estimator provides similar or better accuracy then doing so with synthetic data generated by the \tg. However, the proposed model provides a framework for augmenting a simulator with real-like capabilities and can leverage tedious policy learning processes. 

\ts was shown to embed contact force data within the generated images. Hence, future work could address the calibration of simulated linear forces to real-world contact forces. Future work may also include investigating the integration of \ts based policy learning into robotic in-hand manipulation, addressing the complex real-world challenges and constraints of online learning.

% \balance
\bibliographystyle{IEEEtran}
\bibliography{ref}

\end{document}